\documentclass{article}  
\pdfoutput=1
\usepackage{arxiv}
\usepackage[utf8]{inputenc} 
\usepackage[T1]{fontenc}    
\usepackage[colorlinks=true, allcolors=blue]{hyperref}       
\usepackage{url}            
\usepackage{booktabs}       
\usepackage{amsfonts}       
\usepackage{nicefrac}       
\usepackage{microtype}      
\usepackage{lipsum}		
\usepackage{graphicx}
\usepackage{natbib}
\usepackage{doi}
\usepackage{amsmath}
\usepackage{multirow}
\usepackage{makecell}
\usepackage{subcaption}
\graphicspath{ {.} }

\title{Application of DatasetGAN in medical imaging: preliminary studies}

\author{
  Zong Fan\\
  Department of Bioengineering\\ 
  University of Illinois at Urbana-Champaign\\
  \texttt{zongfan2@illinois.edu}\\
  \And
  Varun Kelkar\\
  Department of Elect. \& Computer Eng. \\
  University of Illinois at Urbana-Champaign \\
  \texttt{vak2@illinois.edu}
  \And 
  Mark A.~Anastasio\\
  Department of Bioengineering\\ 
  University of Illinois at Urbana-Champaign \\
  \texttt{maa@illinois.edu}
  \And
  Hua Li\thanks{Send correspondence to Hua Li. E-mail: huali19@illinois.edu, Telephone: 1 217 326-3122}\\
  Department of Bioengineering\\ 
  University of Illinois at Urbana-Champaign \\ 
  \texttt{huali19@illinois.edu}
  }

\hypersetup{
    pdfsubject={cs.CV, cs.AI},
    pdfauthor={Zong Fan, Varun Kelkar, Mark A.~Anastasio, Hua Li},
    pdfkeywords={Medical Imaging Analysis, GAN, Image Synthesis, Semantic Segmentation},
  }

\begin{document}
\maketitle

\begin{abstract}
\maketitle

Generative adversarial networks (GANs) have been widely investigated for many potential applications in medical imaging.
DatasetGAN is a recently proposed framework based on modern GANs that can synthesize high-quality segmented images while requiring only a small set of annotated training images. 
The synthesized annotated images could be potentially employed for many medical imaging applications, where images with segmentation information are required. However, to the best of our knowledge, there are no published studies focusing on its applications to medical imaging. 
In this work, preliminary studies were conducted to investigate the utility of DatasetGAN in medical imaging.
Three improvements were proposed to the original DatasetGAN framework, considering the unique characteristics of medical images.
The synthesized segmented images by DatasetGAN were visually evaluated. 
The trained DatasetGAN was further analyzed by evaluating the performance of a pre-defined image segmentation technique, 
which was trained by the use of the synthesized datasets.
The effectiveness, concerns, and potential usage of DatasetGAN were discussed. 
\end{abstract}


\section{Introduction}
\label{sec:purpose}

Deep learning is playing an important role in medical imaging applications, ranging from disease diagnosis~\citep{Yadav2019DeepCN} to lesion segmentation~\citep{shorten2019survey}. 
However, training deep neural networks (DNNs) usually needs large amounts of labeled data to achieve satisfying model performance~\citep{shorten2019survey}. 
Annotating a large-scale dataset requires a great deal of time and effort, especially for pixel-level annotation tasks like semantic segmentation. It is particularly true in the medical imaging domain since pixel-level annotation needs a skillful expert in the specific field which is very expensive. 
In addition, it's not easy to generalize a trained model outside the training dataset, especially when the images are captured via different sensors~\citep{Aggarwal2021DiagnosticAO}. 
In this case, the time-consuming re-labeling of the dataset across different imaging sensors is usually required to tune the model but is undesired in practical medical applications~\citep{zhuang2020comprehensive}. 

To reduce the amount of pixel-level annotated images required for training, multiple semi-supervised learning (SSL) methods have been proposed for the semantic segmentation task, which aims to learn with a small labeled dataset augmented by a large unlabeled dataset. 
Some commonly employed methods include pseudo-labeling to generate artificial labels for the unlabeled data~\citep{Lee2013PseudoLabelT}, adversarial learning~\citep{zhou2019collaborative}, consistency regularization~\citep{Ouali2020SemiSupervisedSS}, etc.
Although these methods enable training with small labeled datasets, they may not learn the intrinsic data distribution from the whole dataset with both labeled and unlabeled data, leading to overfitting and hampering model generalization capability~\citep{Li2021SemanticSW}. 

To address the problem by modeling the target data distribution, the generative adversarial network (GAN) is introduced in SSL to augment the training dataset with synthetic data for improving segmentation performance.
GANs are a class of generative approaches that seek to approximate an unknown high-dimensional data distribution by learning to map a sample from a tractable, low dimensional distribution to a sample from the desired data distribution~\citep{Goodfellow2014GenerativeAN}. 
They have been investigated for potential applications in medical imaging, such as image synthesis, image reconstruction and image translation \citep{Kazeminia2020GANsFM}. 
DatasetGAN \citep{Zhang2021DatasetGANEL} is a recently developed SSL semantic segmentation framework that utilizes the information learned by StyleGAN\citep{Karras2019ASG}, a modern GAN which is capable of learning a disentangled representation of the desired data distribution, to generate datasets of semantically segmented images with minimal human intervention. 
By decoding the StyleGAN's pixel-wise feature vectors, DatasetGAN could synthesize both images and their semantic segmentation masks. 
The promise of DatasetGAN comes from the fact that only a small set of annotated images is required to train the decoder since it is able to effectively utilize the semantic information already learned by the underlying StyleGAN.
Therefore, it could produce realistic and unlimited segmented images, which may benefit downstream tasks in medical image analysis utilizing segmentation data.
For example, these segmented images could be used as ground truth to evaluate the performance of a deep learning-based image segmentation algorithm. 

Although DatasetGAN has such attractive properties, its applications on medical imaging remain unexplored. 
In this work, preliminary studies were conducted to investigate its use in medical imaging 
based on a public medical image dataset, SegTHOR~\citep{Lambert2020SegTHORSO}. In addition, three improvements were proposed to the original DatasetGAN framework, considering the unique characteristics of medical images.
The synthesized segmented images by DatasetGAN were visually evaluated. 
The trained DatasetGAN was further analyzed by evaluating the performance of a pre-defined image segmentation technique, 
which was trained by the use of the synthesized datasets.
The effectiveness, concerns, and potential usage of DatasetGAN were discussed as well. 

\vspace{-0.35cm}
\section{Methods}
\label{sec:method}
\subsection{Background of DatasetGAN}
\label{sec:datasetgan}
\begin{figure}[h]
    \centering
    \includegraphics[width=\textwidth]{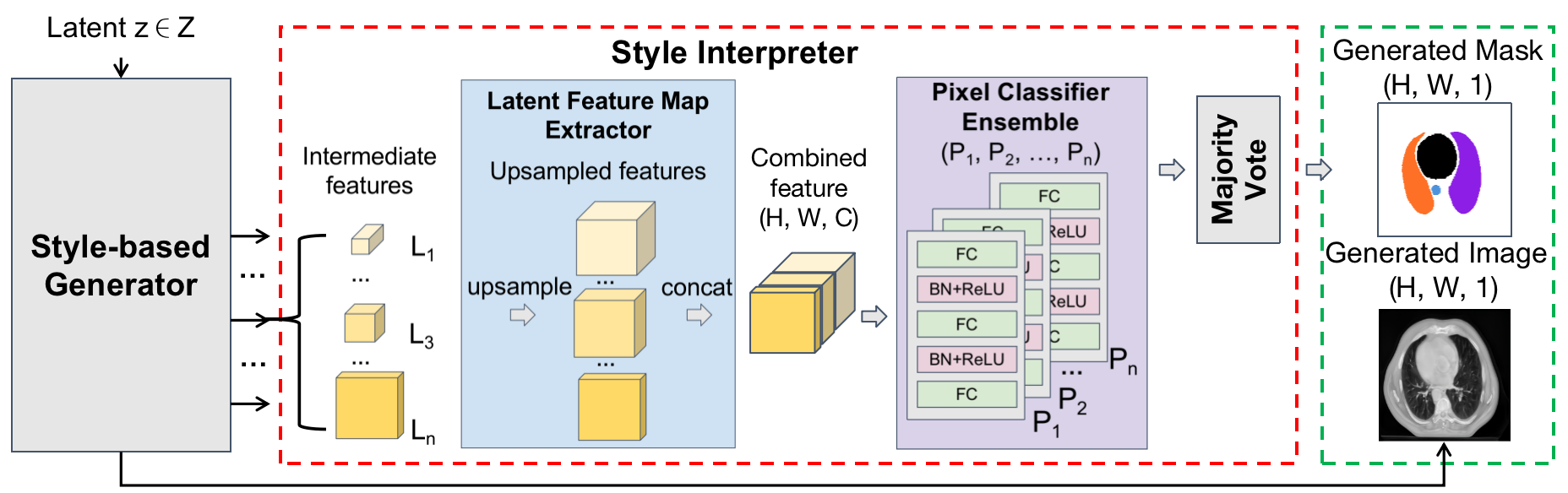}
    \caption{The framework of DatasetGAN to synthesize segmented image.}
    \label{fig:workflow}
    \vspace{-0.4cm}
\end{figure}

As shown in Fig.~\ref{fig:workflow}, the original DatasetGAN architecture is mainly summarized as two parts -- (1) the style-based generator, and (2) the style interpreter. 
DatasetGAN employs the style-based generator, which is identical to the generative network of StyleGAN, to produce images. 
The style-based generator is usually trained on large, unlabelled datasets without data augmentation.
As described in StyleGAN~\citep{Karras2019ASG}, the architecture of the style-based generator contains a mapping network $f$ with $N$ fully-connected layers and a synthesis network $g$ with $M$ progressive style blocks, where $N$ and $M$ are adjustable according to the training image size. The detailed network description could be found in the StyleGAN paper~\citep{Karras2019ASG}. 
Network $f$ maps a latent vector $z$ drawn from a standard normal distribution to an intermediate latent vector $w$, known as a style vector. 
The first style block of $g$ takes $z$ and $w$ as inputs. The rest style blocks progressively take the output feature map of previous style block and a copy of $w$ as inputs, while the final output is the target image with desired resolutions.

The style interpreter employs a latent selection strategy that selects all feature maps from adaptive instance normalization (AdaIN) layers of all style blocks in the previous style-based generator.
The extracted feature maps are up-sampled and concatenated via the latent feature map extractor to produce a feature tensor $F$ with the same size as the synthetic image generated by the style-based generator. 
Next, a pixel classifier ensemble, consisting of multiple classifiers with identical architectures, predicts the class label of a certain pixel on top of the latent feature map $F$ via majority voting. 
By iterating over all pixels on $F$, the segmented image corresponding to the synthetic image is obtained. 

\subsection{Revised DatasetGAN architecture for medical images}
\label{sec:datasetgan-r}
\begin{figure}[!ht]
    \centering
     \includegraphics[width=0.75\textwidth]{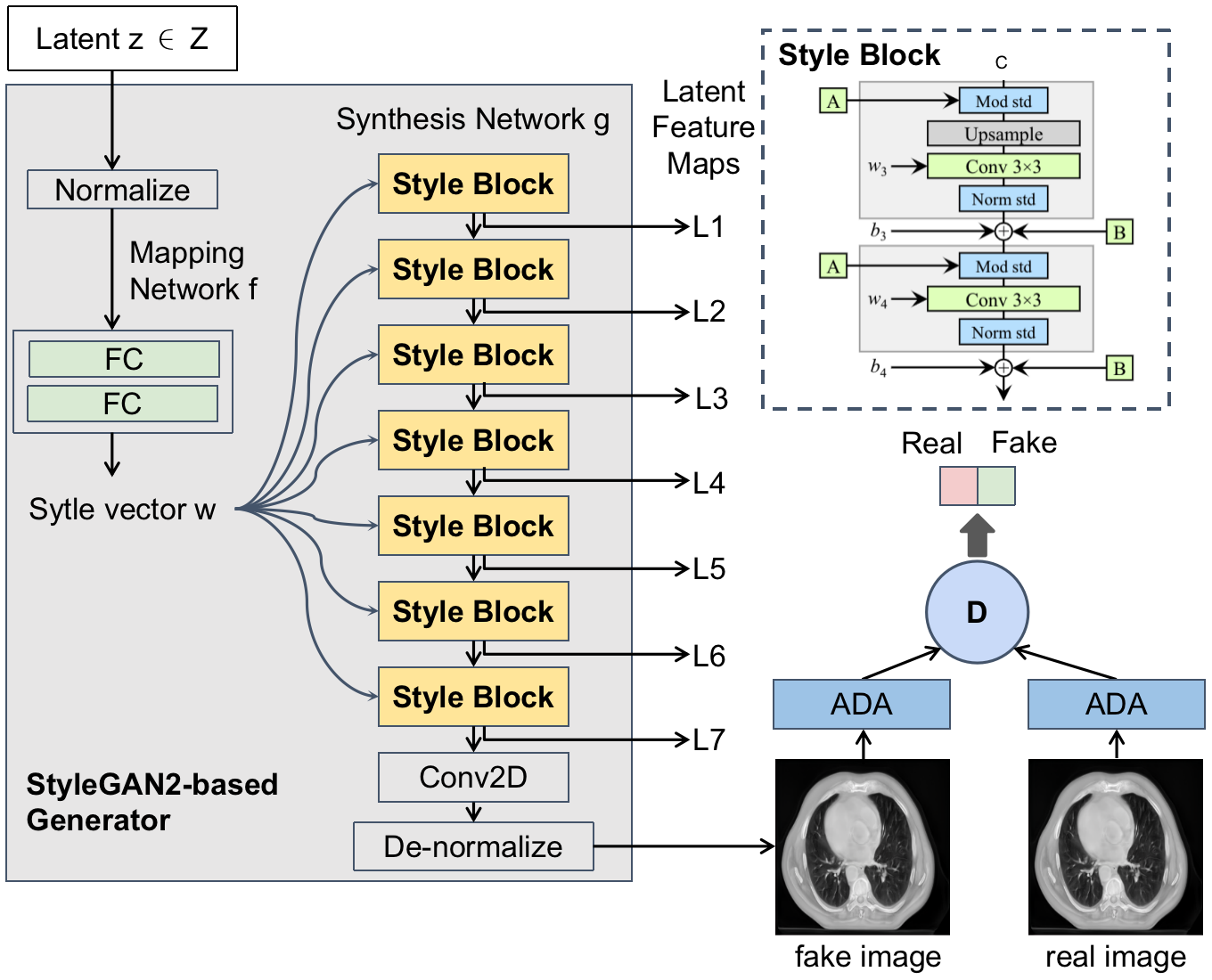}
     \caption{The architecture of revised StyleGAN2-based generator in DatasetGAN-R. 
    The style block (top right corner) of StyleGAN2 is proposed by Karras et.al in the StyleGAN2 paper~\citep{Karras2020AnalyzingAI}.}
     \label{fig:stylegan2}
\end{figure}

Based on the characteristics of medical image datasets, three revisions was developed to the implementation of DatasetGAN in this work as shown in Fig.~\ref{fig:stylegan2}. 
The revised DatasetGAN is referred to as DatasetGAN-R in the rest of the paper.

\noindent \textbf{Revision 1: substituting StyleGAN2 for StyleGAN.} Training a high-performance style-based generator is essential to generate high-quality images simulating the sought-after data distribution. 
Although StyleGAN used in original DatasetGAN can produce visually realistic images, they have been shown to exhibit artifacts~\citep{Zhang2021DatasetGANEL}. 
Several improvements have been proposed to the StyleGAN architecture and training strategy in StyleGAN2~\citep{Karras2020AnalyzingAI}, leading to the improvement of generative performance in terms of commonly used evaluation metrics. More details about these improvements can be found in the StyleGAN2 paper~\citep{Karras2020AnalyzingAI}.

\noindent \textbf{Revision 2: applying adaptive discriminator augmentation (ADA) strategy to training StyleGAN2.} 
Training a high-quality, high-resolution GAN generator usually needs $10^5$ to $10^6$ images to avoid overfitting~\citep{karras2020training}. 
For example, the original DatasetGAN used NABirds dataset with 48k images to train StyleGAN without data augmentation~\citep{Zhang2021DatasetGANEL}. 
Although a few large medical image datasets are available, broad application of DatasetGAN to medical imaging requires that the style-based generator must learn from smaller datasets that are much more common and easier to collect in practical situations. 
To alleviate the overfitting problem on small datasets, ADA proposed by Karras et.al~\citep{karras2020training} is employed before discriminator as data augmentation when training the style-based generator. 
It is a specially designed data augmentation pipeline that tunes the augmentation strength dynamically based on the degree of overfitting, which enables training the discriminator without leaking the augmentation pattern. The experiment results indicate that ADA could significantly improve the performance of style-based generator on small datasets~\citep{karras2020training}.

\noindent \textbf{Revision 3: employing revised latent selection strategy of StyleGAN2.}
The original latent selection strategy selects latent feature maps from two consecutive AdaIN layers of each style block (see right upper corner of Fig.~\ref{fig:stylegan2}) in the style-based generator. 
It may produce an extremely large feature tensor when the input image has very large resolutions. 
In this case, it is probably infeasible to be loaded on a single GPU card, making the training procedure complicated and slow. 
For example, up to 14 latent feature maps are extracted from an image with $256\times 256$ pixels, leading to a high-dimensional feature tensor with $256\times 256\times 4992$ dimensions.  
It's very likely that these densely extracted latent feature maps might contain redundant information for decoding the semantic information of the pixels. 
Therefore, instead of AdaIN layer, the output layer of each style block was selected to produce the latent, which significantly reduces the dimension of the latent feature map by half. 

\vspace{-0.2cm}
\section{Datasets and implementation details}
\label{sec:impl}
\subsection{Datasets}
\label{sec:dataset}
\begin{figure}[!ht]
  \centering 
  \includegraphics[width=0.75\textwidth]{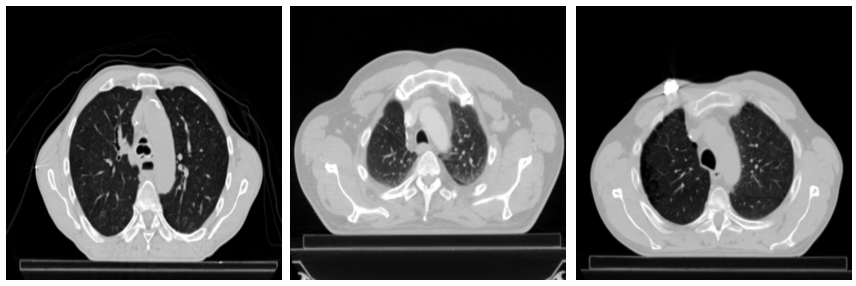}
  \caption{Examples of SegTHOR CT slices.}
  \label{fig:real_data}
\end{figure}

In this study, a computed tomography (CT) dataset called SegTHOR~\citep{Lambert2020SegTHORSO} was employed to train the style-based generator. 
This dataset contains 11084 slices with the image size as $512\times 512$ pixels from 60 patients with non-small cell lung cancer. 
2377 slices with clear lungs in the CT images but without tumor embedded were selected as the training dataset.
Fig.~\ref{fig:real_data} shows several training examples. 

\subsection{Training style-based generator in DatasetGAN-R}
\label{sec:train_gan}
The official stylegan2 with ADA strategy repository 
was employed to train the StyleGAN2-based generator (\url{https://github.com/NVlabs/stylegan2-ada-pytorch}). 
In our study, the number of fully connected (FC) layers in the mapping network was set to $2$ and the number of style blocks in the synthesis network was $7$, producing an image with $256\times256$ pixels. 
The SegTHOR training images were preprocessed by center-cropping and resizing to fixed image size as $256\times 256$ pixels. 
The ADA strategy was employed before the discriminative network of the StyleGAN2 framework when training the style-based generator. 
The Adam stochastic gradient algorithm~\citep{Kingma2015AdamAM} was employed as the optimization algorithm to update the network parameters, with decay rate $\beta_1=0,\beta_2=0.99$ and initial learning rate $lr=0.0025$. 
The training was finished after learning from the dataset for 400 epochs.

\subsection{Training revised style interpreter in DatasetGAN-R}

\label{sec:train_inter}
The style interpreter was trained based on a well-trained StyleGAN2 generator. 
To train the style-interpreter, the first step was to generate the training dataset for the pixel classifier ensemble. 
$N_{si}$ random latent vectors were input into the trained StyleGAN2 generator to synthesize $N_{si}$ images and related latent feature maps. 
Then these images were manually annotated to get the ground-truth (GT) segmentation masks with contours of five organs, including heart, aorta, trachea, left-lung, and right-lung. 
In this study, $N_{si}=30$ was set as the basic training dataset to evaluate the performance of DatasetGAN-R. 

In the second step, our feature extraction strategy extracted 7 feature maps from the StyleGAN2 generator which were defined as $\{L_1, L_2, ..., L_7\}$ with progressively increasing feature map size from $4\times 4$, and $8\times 8$, to $256\times 256$ pixels. 
The 7 feature maps were up-sampled to $256\times 256$ pixels and concatenated into a feature tensor $F$ via latent feature map extractor. 
Each 3D feature tensor had shape of $(256, 256, C)$, where $C=\sum C_i$ and $C_i$ was the number of channels of each feature $L_i$. Particularly, $C$ is 2496 in this study.  
The latent feature maps and corresponding segmentation masks were flattened to produce a pixel classifier training dataset with $256\times256\times N_{si}$ pairs of 2496-D input pixel vectors and 1-D GT pixel labels. 

Next, the pixel classifier ensemble was trained with the prepared dataset. 
The number of classifiers in the pixel classifier ensemble was set to $10$ and the network architecture was the same as that used in the original DatasetGAN. 
Each classifier contained 3 FC layers interconnected with a batch normalization (BN) layer and a rectified linear activation (ReLU) layer, and each FC layer had 2496, 256, 128 neurons, respectively. 
During training, each classifier in the ensemble was iteratively trained on the same dataset while the parameters of each network were initialized independently. The classifier was trained with cross-entropy loss for 100 epochs, and Adam~\citep{kingma2014adam} optimizer was employed with the initial learning rate as 0.0001.

\subsection{Performance evaluation metrics}
\label{sec:metrics}
To evaluate the quality of the synthetic images from the style-based generator, Frechet Inception Distance (FID) was employed which is a widely used GAN evaluation metric~\citep{Heusel2017GANsTB}. 
To qualify the generated segmentation masks, 
mis-classified ratio (m-ratio) was introduced as a metric by comparing the synthetic segmentation mask and the GT segmentation mask corresponding to the same synthetic image, 
defined as the ratio of the mis-classified pixels to the total number of pixels of all classes on the image.


Basically, if the synthetic dataset is decent enough, the segmentation network trained on it should have similar performance as being trained on a real dataset.
Therefore, to further quantify the performance of DatasetGAN, the synthetic segmented image dataset was employed to train a pre-defined segmentation network and the network's segmentation performance was used as the indirect indicator.
As shown in Eq.~\ref{eq:iou} Mean intersection over union (mIoU) was employed to evaluate the performance of the trained segmentation network by comparing the predicted segmentation masks to the GT ones which were annotated manually. 

\begin{equation}
  mIoU=\cfrac{\sum_i TP_i/(TP_i+FP_i+FN_i)}{n}
\label{eq:iou}
\end{equation}
where $TP_i$ is the number of correctly classified (true positive) pixels of class $i$, $FP_i$ is the number of wrongly classified (false positive) pixels of class $i$, $FN_i$ is the number of wrongly misclassified pixels of class $i$, and $n$ is the number of categories. 

\vspace{-0.4cm}
\section{Results}
\label{sec:results}

\vspace{-0.1cm}
\subsection{Quality of segmented image synthesized by DatasetGAN-R}
\label{sec:datasetgan_gen_data}
\begin{figure}[h]
  \centering
  \includegraphics[width=0.8\textwidth]{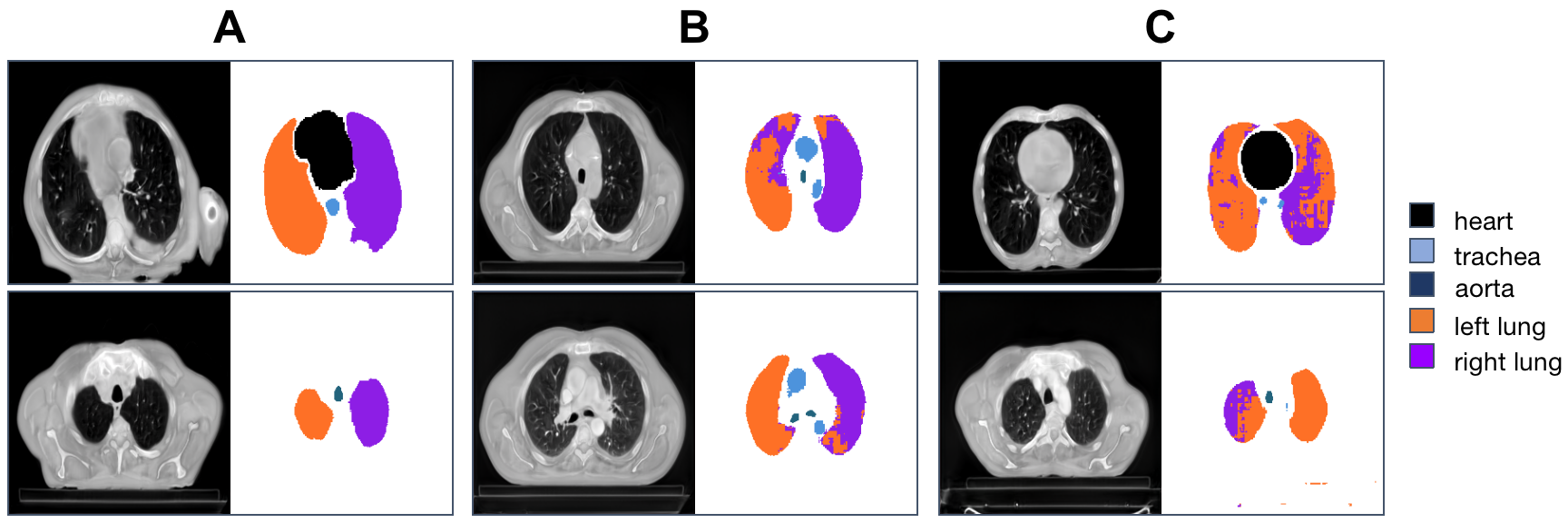}
  \caption{Examples of DatasetGAN generated images and segmentation masks. \textbf{A}: good case; \textbf{B}: partial-bad case; \textbf{C}: bad case.}
  \label{fig:generated_data} 
\end{figure}

One hundred random Latents $zs_{100}$ was used as the input of DatasetGAN-R to synthesize 100 pairs of images and segmentation masks.
The generated images were manually labeled to obtain the pixel-level semantic segmentation masks, used as the GT labels. 
After computing the m-ratio defined in section ~\ref{sec:metrics}, the quality of segmented masks can be grouped into 3 categories: good (m-ratio$<1\%$), partial-bad ($1\%<$m-ratio$<25\%$), and bad image(m-ratio$>25\%$). 
The ratio between the good, partial-bad, and bad images was $74\%$, $15\%$, to $11\%$, respectively. 
Segmented image and mask examples of these 3 categories were shown in Fig.~\ref{fig:generated_data}.
This qualitative analysis result indicates that the majority of the synthesized segmented images are considerably good but still have noises. 

Intuitively, adding more annotated images to train the style interpreter is very likely to improve the interpreter's performance, thus boosting the overall performance of DatasetGAN-R with reduced noises in the synthetic segmentation masks.
For comparison, the number of the annotated images $N_{si}$ for style interpreter training was improved to 100. Then repeat the previous experiment using the same latents $zs_{100}$. 
In this way, the ratio of the three categories was $80\%$, $11\%$, and $9\%$, respectively, indicating that the performance of DatasetGAN-R did get improved with a larger style interpreter training dataset. 

\vspace{-0.2cm}
\subsection{Performance of a pre-defined image segmentation method trained with DatasetGAN-R generated data}
\label{sec:eval}
As discussed in section ~\ref{sec:metrics}, the synthetic segmented images can be employed as the training data of a pre-defined deep-learning image segmentation method. 
Therefore, the performance of the segmentation method is a surrogate of the performance of DatasetGAN-R, 
as described in the study of DatasetGAN~\citep{Zhang2021DatasetGANEL}.

Here, DeepLab-V3-ResNet101~\citep{chen2017rethinking} was used as the segmentation network, 
which was trained by the use of $500$ randomly synthesized image-mask pairs (Train-G). 
To observe the behavior of the network trained on synthetic data or real data, $500$ real image-mask pairs (Train-R) were used as a training dataset for comparison. 
Likewise, two testing datasets, DatasetGAN-R generated testing dataset (Test-G) and real testing dataset (Test-R), were employed to evaluate the trained network.
Test-G contained 100 generated image-mask pairs while Test-R had 100 human-labeled real image-mask pairs. 
The segmentation performance was evaluated by use of 5-fold cross-validation strategy, while the metrics of mIoU and standard deviation (SD) were calculated. 

As shown in Table~\ref{table:mix}, the DeepLab-V3 trained on the synthetic dataset Train-G achieves mIoU of $0.770\pm 0.012$ on Test-G and $0.551\pm 0.011$ on Test-R.
It indicates that there might be an obvious distribution gap between the generated images and real images that causes the large performance inconsistency. 
Similarly, the DeepLab-V3 trained on the real dataset Train-R shows mIoU as $0.459\pm 0.004$ on test-G and $0.799\pm 0.010$ on test-R, showing the preference on real data but still obvious performance gap.
To improve the quality of the synthetic dataset to narrow the performance gap and improve segmentation performance, 3 potential methods were employed and analyzed as below. 

\noindent\textbf{Performance improvement by adding real examples to synthetic dataset}:
The first method was to add small amounts of real data into the synthetic dataset to bridge the distribution gap. 
A mixed training dataset with $500-n$ synthetic images and $n$ real SegTHOR images was made as the training dataset, named as Mix-$n$. 
As shown in Table~\ref{table:mix}, the gap can be narrowed significantly by adding very few real data. Particularly, adding 10 real images could greatly improve the quality of the mixed dataset to train the segmentation network with comparable performance as the real data does. 
It demonstrates that adding few real data, even 5 images, could effectively prevent the intrinsic bias in the synthetic dataset being learned in the following application. 

\vspace{-0.3cm}
\begin{table}[!ht]
  \centering
  \begin{tabular}{c||c|c}
  \hline 
 Type of training dataset & mIoU on Test-G & mIoU on Test-R  \\
  \hline
  Train-G & $0.770\pm 0.012$ & $0.551\pm 0.011$ \\
  Train-R & $0.459\pm0.004$ & $0.799\pm0.010$ \\
  Mix-1 & $0.773\pm 0.011$ & $0.585\pm0.005$ \\
  Mix-5 & $0.769\pm0.012$ & $0.689\pm0.003$ \\
  Mix-10 & $0.768\pm0.011$& $0.755\pm0.004$ \\
  Mix-30 & $\mathbf{0.770\pm 0.011}$ & $\mathbf{0.780\pm 0.003}$ \\
  \hline
\end{tabular}
\captionof{table}{Segmentation model training dataset type vs mIoU. Train-G: 500 synthetic images; Train-R: 500 real images; Mix-n: mixed dataset with $500-n$ generated images and $n$ real images.}
\label{table:mix}
\end{table}
\vspace{-0.3cm}


\noindent \textbf{Performance improvement by increasing the pixel classifier training dataset size}:
As discussed in Section~\ref{sec:datasetgan_gen_data}, the qualitative analysis result implies the effectiveness of increasing the dataset size for training pixel classifiers in improving the quality of generated images. 
To improve the performance of pixel classifiers, experiments were conducted by increasing the number of annotated images $N_{si}$ in the style interpreter training dataset from 30, 50 to 100, respectively. 
Table~\ref{table:pixel_datasize} shows that the mIoU increases $7.1\%$ and $11.2\%$ on Test-G and Test-R respectively by enlarging the training dataset to 100 images. But the performance on the Test-G is still much higher than that on the Test-R.
It indicates that this method could significantly improve the overall performance of DatasetGAN-R but slightly narrow the distribution gap.

\noindent\textbf{Performance improvement by removing the noisy data}:
The third method was to remove the noisy images in the synthetic dataset which may account for the distribution gap. 
The noisy images were manually removed by filtering out the partial-bad or bad images in the synthetic dataset as denoted in section~\ref{sec:datasetgan_gen_data}. 
As shown in Table~\ref{table:noise}, no matter remove only heavy-noisy images (bad images) or all noisy images (bad images and partial-bad images), there is a marginal improvement. These results show that removing noisy data cannot improve the segmentation performance and fill the distribution gap.
One possible guess is that the small portion of noisy generated data may function like regularization terms, causing a negligible negative impact.

\noindent \begin{minipage}{0.5\textwidth}
  \centering
  \centering
  \resizebox*{\textwidth}{!}{
  \begin{tabular}{c||c|c}
    \hline 
    Dataset size & mIoU on Test-G & mIoU on Test-R  \\
    \hline
    N=30 & $0.773\pm 0.012$ & $0.551\pm 0.006$ \\
    N=50 & $0.798\pm 0.004$ & $0.626\pm 0.004$ \\
    N=100 & $\mathbf{0.844\pm 0.005}$ & $\mathbf{0.663\pm 0.004}$ \\
    \hline
  \end{tabular}
  }
  \captionof{table}{The number of images for training style interpreter vs mIoU. $N$: the number of images in the style  interpreter training dataset.}
  \label{table:pixel_datasize}
\end{minipage}
\begin{minipage}{0.5\textwidth}
  \centering
\resizebox*{\textwidth}{!}{
  \begin{tabular}{c||c|c}
    \hline 
    Dataset type & mIoU on test-G & mIoU on test-R  \\
    \hline
    Raw & $0.769\pm 0.011$ & $0.547\pm 0.004$ \\
    No bad & $0.773\pm 0.011$ & $0.556\pm 0.009$ \\
    Only good & $0.770\pm0.012$ & $0.563\pm 0.015$ \\
    \hline
  \end{tabular}
  }
  \vspace{0.05cm}
  \captionof{table}{Segmentation net training data noise vs mIoU. Raw: good/partial-bad/bad=757/135/108; No Bad: good/partial-bad=854/146; Only good: good=1000. }
  \label{table:noise}
\end{minipage}

\vspace{-0.2cm}
\subsection{Effect of the proposed revisions to the performance of DatasetGAN}
\label{sec:abl_datasetgan}
\noindent \textbf{Using StyleGAN2 generator and ADA strategy}: With these two revisions during training StyleGAN2-based generator, the Frechet inception distance (FID) score of the synthetic images of the StyleGAN2 generator improved to $43.02$, while the FID of generated images using the original DatasetGAN architecture only reached $56.34$.
The result shows that the generator achieves better performance with proposed revisions but still has large space for further improvement. 

\noindent \textbf{Revising latent selection strategy}: 
The success of DatasetGAN demonstrates that latent representations extracted from the style-based generator can be decoded to produce semantic segmentation of the generated image~\citep{Zhang2021DatasetGANEL}.
In StyleGAN-based generator, different style blocks control the different visual attributes of the generated images~\citep{Karras2019ASG}. 
We ablated the extracted feature maps to explore which layers of style blocks contribute most in rendering the semantic information realistically. 
As shown in Table~\ref{table:abl}, the high-level feature seems to have more fine-grained semantic information, which is more important for semantic segmentation tasks. Using top 4 layers $\{L_4-L_7\}$ achieves comparable performance to that of using all 7 layers. 
When using this strategy, the pixel classification process could save up to $66.5\%$ running time (from 0.185s to 0.062s) on an NVIDIA GeForce GTX 1080Ti card. 
It confirms our assumption that the feature maps extracted in the original DatasetGAN contain redundant information. A proper selection strategy could accelerate running speed without performance degradation.

\begin{table}[!ht]
  \centering
  \begin{tabular}{c||c|c}
    \hline 
    Selected features& mIoU on Test-G & mIoU on Test-R  \\
    \hline
    $L_1-L_7$ & $0.773\pm 0.012$ & $0.551\pm 0.006$ \\
    $L_2-L_7$ & $0.772\pm 0.011$ & $0.556\pm 0.007$ \\
    $L_3-L_7$ & $\mathbf{0.775\pm 0.010}$ & $\mathbf{0.562\pm 0.008}$ \\
    $L_4-L_7$ & $0.770\pm 0.012$ & $0.550\pm 0.005$ \\
    $L_5-L_7$ & $0.757\pm 0.013$ & $0.511\pm 0.007$ \\
    $L_6-L_7$ & $0.732\pm 0.018$ & $0.522\pm 0.006$ \\
    $L_7$ & $0.680\pm 0.012$ & $0.411\pm 0.008$ \\
    \hline
  \end{tabular}
  \vspace{0.05cm}
  \captionof{table}{Ablation study of feature map selection vs mIoU. $L_i$: the extracted feature from $i$th style block in the style-based generator.}
  \label{table:abl}
\end{table} 

\vspace{-0.4cm}
\section{Discussion \& Conclusion}
To the best of our knowledge, this is the first study that investigates the potential applications of DatasetGAN in medical imaging field.
The experimental results show that DatasetGAN can be employed to provide reliable segmented medical images to support clinical practice. 
Our revisions on the style-based generator and latent selection strategy are effective to tailor DatasetGAN for small datasets. 
Factors such as training dataset size and latent selection strategy which may have impacts on the performance of DatasetGAN-R are also investigated.
In the future, more studies should be conducted to further improve the quality of synthesized images (e.g., the accuracy of masks) to warrant its potential to support clinical practice.

\section{Acknowledgements}
This work was supported in part by NIH awards R01EB020604, R01EB023045, R01NS102213, R01CA233873,
Cancer Center at Illinois seed grant, and DoD Award No. E01 W81XWH-21-1-0062.

\vspace{-0.4cm}
\normalsize{
\bibliographystyle{unsrtnat}
\bibliography{datasetgan}
}

\end{document}